\documentclass[pdflatex,sn-mathphys]{sn-jnl}

\jyear{2022}%
\UseRawInputEncoding
\theoremstyle{thmstyleone}%
\usepackage{bbm}
\theoremstyle{thmstyletwo}%

\theoremstyle{thmstylethree}%
\usepackage{adjustbox}
\usepackage{commath}
\usepackage{tablefootnote}
\usepackage{xcolor}

\begin{document}

\title[ ]{How To Effectively Train An Ensemble Of Faster R-CNN Object Detectors To Quantify Uncertainty}

\author*[1]{\fnm{Denis Mbey} \sur{Akola}}\email{denis.akola@ip-paris.fr}

\author[2]{\fnm{Gianni} \sur{Franchi}}\email{gianni.franchi@ensta-paris.fr}

\affil*[1]{\orgname{Institut Polytechnique de Paris}, \orgaddress{\street{Palaiseau}, \postcode{91120}, \country{France}}}

\affil[2]{ \orgname{U2IS, ENSTA Paris, Institut Polytechnique de Paris}, \city{Palaiseau}, \postcode{91120}, \country{France}}



\abstract{This paper presents a new approach for training two-stage object detection ensemble models, more specifically, Faster R-CNN models to estimate uncertainty. We propose training one Region Proposal Network(RPN) and multiple Fast R-CNN prediction heads is all you need to build a robust deep ensemble network for estimating uncertainty in object detection. We present this approach and provide experiments to show that this approach is much faster than the naive method of fully training all $n$ models in an ensemble. We also estimate the uncertainty by measuring this ensemble model's Expected Calibration Error (ECE). We then further compare the performance of this model with that of Gaussian  YOLOv3, a variant of YOLOv3 that models uncertainty using predicted bounding box coordinates. The source code is released at \url{https://github.com/Akola-Mbey-Denis/EfficientEnsemble}}





\maketitle
\section{Introduction}
Recent advances in deep learning~\cite{simonyan2014very,https://doi.org/10.48550/arxiv.1512.03385} have dramatically improved neural network accuracy. As such, deep neural networks are now employed in making complex decisions in applications such as object detection, autonomous driving, and medical diagnosis~\cite{https://doi.org/10.48550/arxiv.1909.10155}.
However, in these safety-critical systems, having accurate predictions is not enough to measure neural network performance~\cite{https://doi.org/10.48550/arxiv.1904.04620}. For example, consider an autonomous vehicle that uses a neural network to detect pedestrians and other obstructions. If the detection network cannot confidently predict the presence or absence of immediate obstructions, the car should rely more on the output of other sensors for braking. This implies we should also have information on how likely incorrect those predictions might be. To do this, we need a criterion to determine how uncertain the predictions are \cite{https://doi.org/10.48550/arxiv.1706.04599}.
The measure of uncertainty in deep neural networks is important in safety-critical systems that rely on object detection models because unchecked or erroneous detections could lead to fatal accidents~\cite{https://doi.org/10.48550/arxiv.1909.10155,https://doi.org/10.48550/arxiv.1904.04620}. 
Also, a valid critique of deep learning-based methods is their black-box nature; since all parts of the problem are learned from data, there is no strict understanding of how the problem is solved by the network~\cite{https://doi.org/10.48550/arxiv.1802.07095}. For real-world applications, it will be helpful to determine how reliable the underlying estimates of deep learning models are before deploying them to avoid or minimize catastrophes~\cite{https://doi.org/10.48550/arxiv.1909.10155,https://doi.org/10.48550/arxiv.1904.04620}.\\ 
To measure uncertainty, various techniques have been employed to calibrate and model uncertainty in object detection models. In this paper, we provide a review of the various techniques used to estimate uncertainty and calibrate objection detection models. \\
Our main contributions are as follows
\begin{enumerate}
    \item We present a new technique to train an ensemble of deep neural networks specifically, an ensemble of Faster R-CNN~\cite{https://doi.org/10.48550/arxiv.1506.01497}  models using only one Region Proposal Network (RPN) and multiple Fast R-CNN prediction heads which we call EfficientEnsemble.
  
      \item We  conduct extensive experiments to compare EfficientEnsemble with other methods including Monte Carlo Dropout~\cite{https://doi.org/10.48550/arxiv.1506.02142} and the traditional Deep Ensembles. 
     These experiments are conducted to identify effective strategies for merging the results from Deep Ensembles. Additionally, we report the empirical variance associated with the output from our ensemble network to measure the discrepancy between merged bounding boxes.
\end{enumerate}
\section{Related Work}
In this section, we provide a review of the current state-of-the-art object detection models, current uncertainty estimation methods, and techniques for calibrating deep neural methods.
\subsection{Object Detection Models}

Object detection is a fundamental task in computer vision and it is the starting point of many computer vision applications such as pedestrian detection in autonomous driving, security, and surveillance~\cite{Jiao2019ASO}. With the development of Convolutional Neural Networks, a lot of progress has been made in this area and deep object detection models have achieved impressive results on various object detection tasks~\cite{https://doi.org/10.48550/arxiv.1506.01497,https://doi.org/10.48550/arxiv.1804.02767}.
\\The widely used nomenclature for classifying deep learning-based object detection models is the number of stages used in the object detection model. Based on this convention, object detection models are categorized as one-stage and two-stage object detection models~\cite{article01}.
In two-stage object detection models, the first stage proposes approximate object regions using deep features. The second stage performs bounding box regression and classification of these object candidate regions~\cite{Sultana_2020}. In one-stage object detectors, classification and regression are done in a single-shot fashion using regular and dense sampling with respect to location, scale, and aspect ratio. Two-stage object detection models have higher localization and recognition accuracy than one-stage models. On the other hand, one-stage models have higher inference speeds than their two-stage counterparts~\cite{https://doi.org/10.48550/arxiv.1804.02767}.
The state-of-the-art two-stage object detection algorithm is the Region Convolutional Neural Network (R-CNN) family of object detectors. These include R-CNN~\cite{DBLP:journals/corr/GirshickDDM13}, Fast R-CNN~\cite{DBLP:journals/corr/Girshick15}, Faster R-CNN~\cite{https://doi.org/10.48550/arxiv.1506.01497}, and Mask R-CNN~\cite{https://doi.org/10.48550/arxiv.1703.06870}. The evolution of two-stage object detection models is attributed to making efficient computations by sharing computations across the convolutional layers in the Region proposal network and the R-CNN layers which resulted in Fast R-CNN~\cite{DBLP:journals/corr/Girshick15}. The selective search method~\cite{DBLP:journals/corr/Girshick15,article01} used for generating approximate object regions in R-CNN~\cite{DBLP:journals/corr/GirshickDDM13} and Fast R-CNN~\cite{DBLP:journals/corr/Girshick15} was replaced by an efficient Region Proposal Network for proposing approximate object regions and this gave birth to Faster-RCNN~\cite{https://doi.org/10.48550/arxiv.1506.01497}. Mask R-CNN is an extension of Faster R-CNN~\cite{https://doi.org/10.48550/arxiv.1506.01497}  model for both object detection and semantic segmentation.\\
The one-stage object detection models include You Only Look Once(YOLO)~\cite{https://doi.org/10.48550/arxiv.1506.02640}, YOLOv2~\cite{8100173}, YOLOv3~\cite{https://doi.org/10.48550/arxiv.2107.08430} and Single Shot Multibox Detection (SSD~\cite{44872}).
You Only Look Once(YOLO)~\cite{https://doi.org/10.48550/arxiv.1506.02640} predicts  class probabilities and bounding boxes from the input image using a simple Convolutional Neural Network (CNN). YOLO~\cite{https://doi.org/10.48550/arxiv.1506.02640} divides an input image into a fixed number of grids. Each cell in these grids predicts a fixed number of bounding boxes with confidence scores. Each confidence score is computed by multiplying the probability to detect that object with the intersection over union (IOU) between the predicted and ground truth bounding boxes. The bounding boxes having a class probability above a certain threshold value are selected and used to locate the objects within the image.\\
The other YOLO models follow the same paradigm, but with some design changes made in them to make them more robust.
By using batch normalization layers instead of dropout in YOLO, a high-resolution classifier and multi-scale training on both COCO~\cite{https://doi.org/10.48550/arxiv.1405.0312} and ImageNet~\cite{ILSVRC15} datasets yielded YOLOv2.
Also, replacing the Darknet-19 backbone network with Darknet architecture with 53 convolutional layers gave birth to YOLOv3~\cite{https://doi.org/10.48550/arxiv.2107.08430}.
Other one-stage include the Single Shot MultiBox Detection (SSD) ~\cite{44872} receives an image as input and passes it through multiple convolutional layers  with different filter sizes.\\
Recently, transformers have made strides in the object detection domain. Transform networks have been used for many years in Natural Language Processing (NLP)  and have made a tremendous impact in this area~\cite{zaidi2022survey}. However, it was until recently that the research community exploited the efficacy of transformers in vision tasks. The  work of ~\cite{dosovitskiy2020image} introduced the first transformer model for image classification called the vision transformer. This work opens the door for the use of transformer models in object detection. Two known transformer-based object detection models include the Detection Transformer (DeTR)~\cite{carion2020end}, and the Swin Transformer~\cite{liu2021swin}. The detection Transformer uses a CNN backbone for feature extraction. The encoder of the transformer takes image features along with position encodings as input and passes its output as input  to the  transformer decoder.  The decoder manipulates the input embeddings, called object queries to generate an output that is passed to multi-layer perceptrons to predict class and bounding boxes.  Swin Transformer~\cite{liu2021swin} provides a computer vision-based transformer backbone. Swin Transformer~\cite{liu2021swin} first splits an input image into multiple, non-overlapping patches and converts them into embeddings. It then applies multiple Swin Transformer blocks in 4 stages with each stage reducing the number of patches to maintain a hierarchical representation.
\vspace{-0.2cm}
\subsection{Uncertainty Estimation Methods}
Uncertainty estimation is an essential requirement in safety-critical systems because it is necessary to know when we cannot trust a model's predictions~\cite{https://doi.org/10.48550/arxiv.1904.04620}. The knowledge of uncertainty helps us to refrain  from taking actions based on predictions when the uncertainty is high. Uncertainty could be aleatoric or epistemic~\cite{DBLP:journals/corr/abs-2107-03342,Abdar_2021}. Aleatoric uncertainty is the uncertainty arising from noisy data~\cite{DBLP:journals/corr/abs-2107-03342,Abdar_2021}. Epistemic uncertainty is the uncertainty arising from a noisy model~\cite{DBLP:journals/corr/abs-2107-03342,Abdar_2021}. Different methods have been exploited to measure uncertainty in deep learning models namely; Bayesian neural networks (BNN)~\cite{DBLP:journals/corr/abs-2107-03342}, deep ensembles~\cite{https://doi.org/10.48550/arxiv.2104.02395}, and Gaussian processes~\cite{pmlr-v31-damianou13a}.
\\Bayesian neural networks~\cite{DBLP:journals/corr/abs-2107-03342} is a class of neural networks that are not just good function approximators but overcome the common problem of neural networks which is over-fitting on training data.  Bayesian methods  provide rich probabilistic interpretations for predicted outcomes via their posterior distributions~\cite{Abdar_2021}. Bayesian inference replaces a point estimate of the weights and its corresponding prediction function from standard training of neural networks with inferred probabilistic distributions of all model parameters and predictions. As a result, uncertainty can be quantified for example computing the variance between model parameters and predictions~\cite{Abdar_2021,DBLP:journals/corr/abs-2107-03342}. However, computing the exact posterior inference is intractable, but techniques have been devised to approximate it. The common approximations include Monte-Carlo dropout~\cite{https://doi.org/10.48550/arxiv.1506.02142} and variational inference~\cite{https://doi.org/10.48550/arxiv.2002.02655}. Monte-Carlo dropout uses dropout to compute prediction uncertainty~\cite{https://doi.org/10.48550/arxiv.1506.02142}. Variational inference (VI) is an approximation method that learns the posterior distribution over BNN weights. Variational inference methods consider the Bayesian inference problem as an optimization problem that neural network optimizers like  stochastic gradient descent(SDG) try to optimize during the  training of deep neural networks (DNN)~\cite{https://doi.org/10.48550/arxiv.2002.02655}.
\\Deep Ensembles is another method for quantifying uncertainty in deep neural networks (DNN). This has been extensively used in many real-world applications~\cite{Hu_2019}. Deep Ensembles are an effective technique that can be used to enhance the predictive performance of supervised learning problems. They obtain better predictions on test data and produce model uncertainty estimates from learners when provided with out-of-distribution data~\cite{Abdar_2021}. To get final prediction outputs in deep ensemble networks, some approaches for merging the predictions of each model in an ensemble network have been useful. Some of the techniques include averaging, majority voting, stacked generalization, and super learning\cite{https://doi.org/10.48550/arxiv.2104.02395}.
\\Deep Gaussian process (GP) models are effective multi-layer decision-making tools that can accurately model uncertainty~\cite{pmlr-v31-damianou13a}. Deep Gaussian processes are a multi-layer hierarchy of Gaussian processes (GPs)~\cite{Rasmussen2004}. GPs models compute  the similarity between data points using a kernel function and use  the Bayes rule to model a distribution over functions by maximizing the marginal likelihood. GPs do not scale well because they require full datasets at inference time  and do not work well when the dimensionality of data increases~\cite{Rasmussen2004}.
\subsection{Calibration Techniques}
The calibration of deep learning models is important because it provides additional information about the prediction confidence of deep learning models. The confidence of deep learning models which we obtain by calibration is a good indicator of the model's trust and helps us avoid fatal accidents in the applications of these deep learning models such as in autonomous driving and medical applications~\cite{https://doi.org/10.48550/arxiv.1706.04599}.  Some methods for  calibrating deep learning models include Platt scaling~\cite{Platt99probabilisticoutputs}, and temperature scaling~\cite{Platt99probabilisticoutputs}.
\\Platt scaling~\cite{Platt99probabilisticoutputs} fits a logistic regression to a classification model. This is usually done on the validation dataset which is used to compute calibrated predictions at test time. This method learns scalar parameters on the held-out validation set and then computes the calibrated probabilities given the uncalibrated logits vector on the test set. Negative log livelihood minimization is a common way of estimating the scalar parameters~\cite{info:doi/10.2196/medinform.3445,https://doi.org/10.48550/arxiv.1904.01685}.
\\Temperature scaling is an extension of Platt scaling~\cite{Platt99probabilisticoutputs}. In temperature scaling, the temperature of the softmax function is often optimized. 
\\Other calibration methods include histogram binning, and isotonic regression~\cite{https://doi.org/10.48550/arxiv.1904.01685}.

\section{Method}
Two-stage detection models, like Faster R-CNN, perform the task of object detection in two stages. As identified earlier, the first stage generates proposals, and the second stage performs bounding box regression and object classification. The second stage of Faster R-CNN embeds Fast R-CNN for the task of bounding box regression and object classification.
Following this structure, we were inspired to perform deep ensemble via bounding box regression using Fast R-CNN in the second stage of Faster R-CNN. The advantage of performing an ensemble using this method is it reduces the computation time in performing three forward passes through the three Faster R-CNN networks and we call our method EfficientEnsemble.
In EfficientEnsemble, all you need is to train one RPN network and three different Fast R-CNN prediction heads. The first model in the ensemble performs a full forward pass through both the RPN and Fast R-CNN stages.
The remaining models in our ensemble take the proposals from the first model and finally regress and classify them. We then used various decision fusion criteria like Soft NMS~\cite{DBLP:journals/corr/BodlaSCD17}  and Weighted Boxes Fusion~\cite{DBLP:journals/corr/abs-1910-13302} to produce the final predictions of our ensemble. EfficientEnsemble is illustrated in Figure ~\ref{fig:proposal}. The region proposals produced by $\texttt{RPN}$ are given as input to  $\texttt{Fast RCNN-1}$,  $\texttt{Fast RCNN-2}$ and $\texttt{Fast RCNN-3}$ prediction heads.  $\texttt{Fast RCNN-1}$ $\texttt{Fast R-CNN-2} $ and $\texttt{Fast R-CNN-3}$ produces refined bounding boxes with their corresponding scores. As seen in Figure ~\ref{fig:proposal}, the outputs of all the Fast R-NN networks from all three models are fused to generate the final predictions of the model.
\begin{figure*}
    \centering
    \includegraphics[scale=0.45]{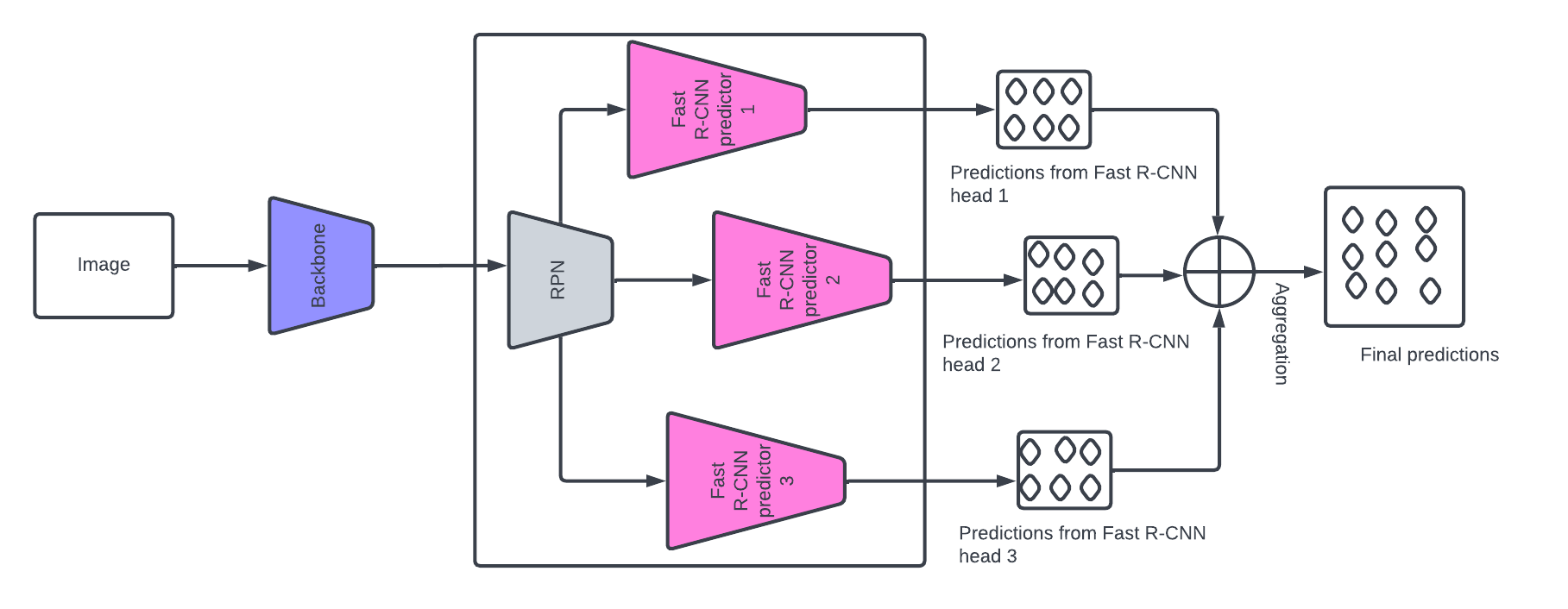}
    \caption{EfficientEnsemble for building Deep Ensembles of  Faster R-CNN models}
    \label{fig:proposal}
\end{figure*}
The traditional approach for ensembling involves using fully trained models. For the sake of comparison, we also built a Deep Ensemble network based on this scheme. Inference in this scheme involves three full forward passes as each of the models in our network performs a forward pass on the input.
Since each model in the ensemble provides predictions for a given input image, we adopted various strategies to fuse the predictions which are discussed in section ~\ref{output_merging}.

\begin{figure*}
    \centering
    \includegraphics[scale=0.45]{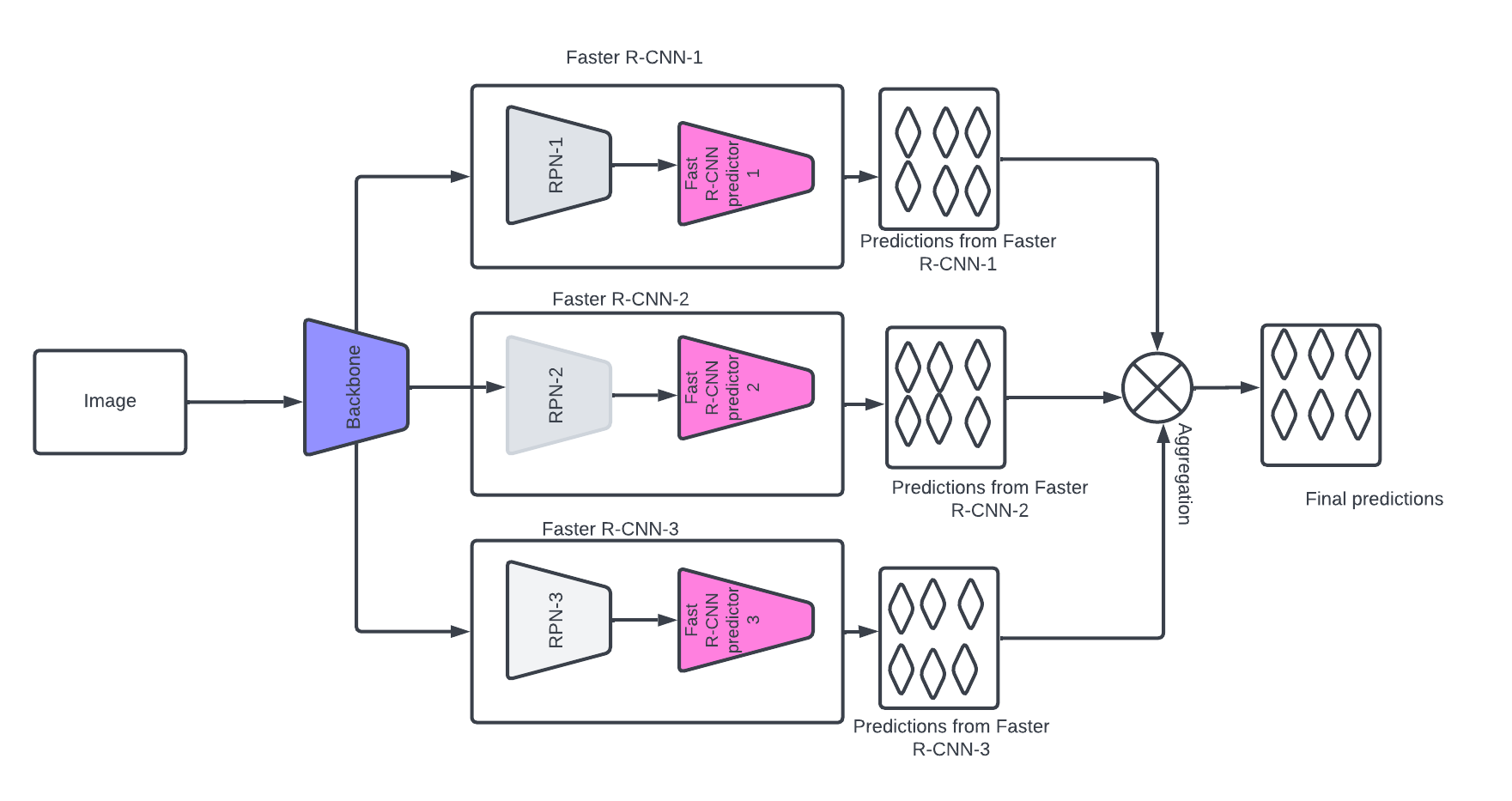}
    \caption{Deep Ensembles using fully-trained Faster R-CNN models}
    \label{fig:proposal}
\end{figure*}
\subsection{Decision Fusion Strategies}
\label{output_merging}
Ensemble learning trains several base learners and aggregates the outputs of base learners
using some decision rules. The output of an ensemble network is often merged using different criteria to yield the final predictions. The aggregating function used to combine the outputs determines the effective performance
of an ensemble. In effect, we adopted three different methods for merging the outputs of our ensemble. \\
The first method uses the Non-Maximum Suppression (NMS) algorithm  to  select  bounding boxes with greater scores from  many overlapping bounding boxes retrieved from all the Deep Ensemble models. Thus, NMS removes spurious bounding boxes. We used an NMS  threshold of 0.50 throughout our experiments.
The second method applies the Soft Non-Maximum Suppression (Soft NMS)~\cite{DBLP:journals/corr/BodlaSCD17} algorithm on the sets of predictions from all models in our ensemble and then yields the final predictions based on a given IOU. Soft NMS works better than the Non-Maximum Suppression (NMS) algorithm because it reduces the confidence scores of predicted bounding boxes as a function of the  intersection over Union (IOU) overlap instead of setting the confidence scores of overlapping bounding boxes to zero in NMS which completely eliminates such predictions. The Soft NMS paper results showed that using this approach improved the performance of object detection models on various object detection tasks.  \\
The second method we leveraged was Weighted Boxes Fusion (WBF) which was inspired by ~\cite{DBLP:journals/corr/abs-1910-13302}. In this approach, the confidence scores of all proposed bounding boxes are used to construct the average boxes which prove more useful than NMS and Soft NMS which simply discards some predictions based on an overlap threshold. WBF algorithm works by  iteratively looping through the set of predicted bounding boxes, finding and forming matching boxes sets based on an IOU threshold, and finally aggregates these boxes in a manner proportionate to the confidence scores of each of the boxes in the matching sets.

\section{Experiments}
\subsection{Datasets}
In this section, we describe the datasets we used for training and evaluation of our Deep Ensembles method. We used MOT17Det~\cite{https://doi.org/10.48550/arxiv.1603.00831}, and KITTI~\cite{Geiger2012CVPR} datasets.
MOT17Det~\cite{https://doi.org/10.48550/arxiv.1603.00831} is a dataset for people detection from the Multi-Object Tracking (MOT) Challenge~\footnote{https://motchallenge.net/data/MOT17Det/}. It comprises of 14 videos under different lighting, view, and weather conditions. 7 videos are used as training data and the last 7 for testing purposes. The MOT16 training dataset was used as validation as the labels of the MOT17DET dataset are not public.
KITTI~\cite{Geiger2012CVPR} 2D object detection dataset contains 7481 training images and 7518 test images with corresponding training labels for the training images. It was collected while driving in and around Karlsruhe, Germany. The 2D Object detection dataset does not have a validation dataset, so we split the training dataset and used 80\% as training data and 20\% as the validation dataset.
BDD100K ~\cite{yu2018bdd100k} is a diverse  driving dataset. The dataset possesses geographic, environmental, and weather diversity and encourages models to learn to generalize better to new conditions. The BDD100K has three parts, a training set with 70K images, a validation set with 10K images, and a test set with 20K images. Labels are not provided for the testing dataset since it is not publically available. Thus, we evaluated our model using on validation dataset.

\subsection{Training Protocol for Deep Ensembles}
We trained an ensemble of Faster R-CNN networks  on MOT17Det~\cite{https://doi.org/10.48550/arxiv.1603.00831}, KITTI~\cite{Geiger2012CVPR} and BDD100K datasets~\cite{yu2018bdd100k}. 
We used the same pre-trained RPN~\cite{https://doi.org/10.48550/arxiv.1506.01497} network for all Faster R-CNN networks. All models were trained on a single NVIDIA GeForce GTX TITAN X GPU. For the training of the models, we adopted different approaches for training the ensemble models. On the first approach we optimize all trainable layers in the backbone, RPN, and Fast R-CNN predictor for all three models in our ensemble. The second method consists in training one RPN layer and three Fast R-CNN predictor layers for each of the three Faster R-CNN models. For the Gaussian YOLO ~\cite{https://doi.org/10.48550/arxiv.1904.04620} model, we trained all trainable layers of the model on the selected datasets.
For all Faster R-CNN models, the backbone network we used was Resnet101~\cite{DBLP:journals/corr/HeZRS15} with Stochastic Gradient Descent with a momentum of 0.90 as an optimizer and a batch size of 4. The learning rate was set to 0.001 for the MOT17DET dataset and 0.00125 for the KITTI dataset.  All Gaussian YOLOv3 models were also trained using a batch size of 4, and with Stochastic Gradient Descent with a momentum of 0.90 as an optimizer with a learning rate of 0.00125 using Darknet-53 as the backbone.
\subsection{Metrics}
Different metrics have been developed to quantify the measure the uncertainty of object detection models. Evaluating the quality of calibration involves measuring the statistical consistency between the predictive distribution and observations~\cite{DBLP:journals/corr/abs-2107-03342}. For the classification task, the quality of uncertainty is estimated using Expected Calibration Error (ECE) \cite{DBLP:journals/corr/abs-2107-03342}, which is computed from the average bin confidence and accuracy.
To compute the average bin confidence, the predictions are ordered by prediction confidence $\hat{p_{i}}$ and grouped into $M$ bins, $b_{1},b_{2},.....b_{M}$~\cite{https://doi.org/10.48550/arxiv.1706.04599}. \\The average bin confidence is computed as 
\begin{equation}
conf(b_{m}) = \frac{1}{\abs{b_{m}}} \sum_{s\in b_{m}} \hat{p_{s}}
\end{equation}
The average bin accuracy is likewise computed as \begin{equation}
    acc(b_{m}) = \frac{1}{\abs{b_{m}}}\sum_{s\in b_{m}} \mathbbm{1} (\hat{y_{s}}-y_{s})
\end{equation}
where $\hat{y_{s}}$, $y_{s}$ and $\hat{p_{s}}$ refers to the predicted label, true class
label, and prediction confidence of a sample $s$, respectively. The average confidences are well-
calibrated when for each bin $acc(b_{m}) = conf(b_{m})$~\cite{https://doi.org/10.48550/arxiv.1706.04599,DBLP:journals/corr/abs-2107-03342}.
The average bin confidence and bin accuracy are often visualized using reliability diagrams~\cite{https://doi.org/10.48550/arxiv.1909.10155,https://doi.org/10.48550/arxiv.1706.04599}.
\\Expected Calibration Error (ECE) is one of the popular uncertainty measures. To compute ECE, $M$ equally spaced bins ,$b_{1},b_{2},...,b_{m}$ are considered. $b_{m}$ is the set of indices of samples whose confidence fall within the interval $I_{m}=\big [\frac{m-1}{M},\frac{m}{M}\big ]$
Thus, \begin{equation}
    ECE = \sum_{m=1}^{M} \frac{\abs{b_{m}}}{N}\abs{acc(b_{m}) -  conf(b_{m})}
    \end{equation} The uncertainty measure we adopted is ECE implemented in a python package from \cite{https://doi.org/10.48550/arxiv.1909.10155}.

\subsection{Experimental Setup}
In this section, we present the results of the different experiments we conducted. The reported results are based on the average precision (AP)  and the average recall (AR) metrics of object detection models computed at IOU of 0.50 and 0.95 using the official COCO object detection evaluation tool. We also computed the Expected Calibration Error (ECE) metric as a measure of the model's uncertainty. Hereafter, we named the original approach of performing an ensemble using fully-trained deep models as the 'Traditional' approach. 
In Tables ~\ref{MOT-16-ensemble}, ~\ref{Kiiti-ensemble}, and ~\ref{bdd100k-ensemble}, we reported the results of all experiments we conducted which includes results for ensembling with the traditional method, EfficientEnsemble and, Monte-Carlo Dropout. We also report results for Gaussian YOLOv3 to see how EfficientEnsemble compares to it.  Gaussian YOLOv3~\cite{https://doi.org/10.48550/arxiv.1804.02767} is a variant of YOLOv3 that was trained to estimate uncertainty by computing the mean and variance of each predicted bounding box coordinates. The variance of each bounding box coordinate serves as a measure of the uncertainty of the coordinate and the mean represents the predicted bounding box coordinate value.
\begin{table*}

       \centering
   \begin{adjustbox}{width=1.0\textwidth}
  \begin{tabular} {ccccccc}
  \textbf{Model type}&\textbf{Ensemble method}&\textbf{Aggregation method}&\textbf{AP} ($\uparrow$) &\textbf{AR} ($\uparrow$) &\textbf{ECE} ($\downarrow$)&\textbf{Eval. Time} ($\downarrow$)\\\hline
  Baseline&None&None&0.749&0.778&0.147&1417.243\\
   Gaussian YOLOv3 &None &None &0.751&0.792&0.154&  1864.397\\
  Deep Ensembles &Traditional&WBF&\textbf{0.773}&\textbf{0.802}&\textbf{0.121}&5653.819\\
  Deep Ensembles &Traditional&NMS&0.773&0.801&0.176&5667.085\\
  Deep Ensembles&Traditional&Soft NMS&0.750&0.770&0.139&5601.562\\
 Deep Ensembles&EfficientEnsemble (ours) &WBF& \textbf{\textcolor{red}{0.770}}&\textbf{\textcolor{red}{0.799}}&\textbf{\textcolor{red}{0.124}}& 3530.634\\
 Deep Ensembles &EfficientEnsemble (ours) &NMS&0.770&0.799&0.176&\textbf{\textcolor{red}{3507.331}}\\
~\tablefootnote{We are comparing the evaluation time of the Deep Ensembles} Deep Ensembles &EfficientEnsemble (ours)&Soft NMS&0.750&0.7740&0.136& \textbf{3463.955}\\
 Deep Ensembles&Monte-Carlo Dropout&WBF&0.703&0.739&0.190&4151.497\\
  Deep Ensembles &Monte-Carlo Dropout&NMS&0.685&0.730&0.405&4148.285\\
 Deep Ensembles&Monte-Carlo Dropout&Soft NMS&0.681&0.711&0.386&3931.765\\

 \hline
\end{tabular}
\end{adjustbox}
      \caption{MOT16 Ensemble results}
    \label{MOT-16-ensemble}
\end{table*}

\begin{table*}

       \centering
   \begin{adjustbox}{width=1.0\textwidth}
  \begin{tabular} {ccccccc}
  \textbf{Model type}&\textbf{Ensemble method}&\textbf{Aggregation method}&\textbf{AP} ($\uparrow$) &\textbf{AR} ($\uparrow$) &\textbf{ECE} ($\downarrow$)&\textbf{Eval. Time} ($\downarrow$)\\\hline
  Baseline&None&None&0.440&0.583&0.1902&1256.093\\
Gaussian YOLOv3 & None & None& 0.316&0.406&0.170& 1790.342\\
  Deep Ensembles &Traditional&WBF&\textbf{0.552}&\textbf{0.640}&\textbf{0.178}&4832.7836\\
  Deep Ensembles &Traditional&NMS&0.487&0.585&0.208&4949.156\\
  Deep Ensembles&Traditional&Soft NMS&0.507&0.562&0.186&4685.959\\
 Deep Ensembles&EfficientEnsemble (ours) &WBF&\textbf{\textcolor{red}{0.547}}&\textbf{\textcolor{red}{0.639}}&0.190&\textbf{\textcolor{red}{3063.079}}\\
 Deep Ensembles &EfficientEnsemble (ours) &NMS&0.490&0.589&0.212&3157.701\\
 Deep Ensembles &EfficientEnsemble (ours) &Soft NMS&0.504&0.588&\textbf{\textcolor{red}{0.183}}&\textbf{2920.513}\\
 Deep Ensembles&Monte-Carlo Dropout&WBF&0.216&0.458&0.347&4460.583\\
  Deep Ensembles &Monte-Carlo Dropout&NMS&0.198&0.428&0.669&4176.558\\
 Deep Ensembles&Monte-Carlo Dropout&Soft NMS&0.156&0.248&0.880&3379.068\\

 \hline
\end{tabular}
\end{adjustbox}
      \caption{KITTI Ensemble results}
    \label{Kiiti-ensemble}
\end{table*}

\begin{table}

       \centering
   \begin{adjustbox}{width=1.0\textwidth}
  \begin{tabular} {ccccccc}
  \textbf{Model type}&\textbf{Ensemble method}&\textbf{Aggregation method}&\textbf{AP} ($\uparrow$) &\textbf{AR} ($\uparrow$) &\textbf{ECE} ($\downarrow$)&\textbf{Eval. Time} ($\downarrow$)\\\hline
  Baseline&None&None&0.202&0.270&0.214&2974.243\\
   Gaussian YOLOv3 &None &None &0.147&0.219&0.184& 1771.182\\
  Deep Ensembles &Traditional&WBF&\textbf{0.219}&\textbf{0.365}&\textbf{0.105}&12288.995\\
  Deep Ensembles &Traditional&NMS&0.188&0.273&0.227&11677.586\\
  Deep Ensembles&Traditional&Soft NMS&0.197&0.320&0.192&12335.994\\
 Deep Ensembles&EfficientEnsemble (ours) &WBF&\textbf{\textcolor{red}{0.215}}&\textbf{\textcolor{red}{0.354}}&\textbf{\textcolor{red}{0.123}}&8001.135\\
 Deep Ensembles&EfficientEnsemble (ours) &NMS&0.184&0.267&0.239&\textbf{7426.009}\\
 Deep Ensemble &EfficientEnsemble (ours) &Soft NMS&0.193&0.309&0.198&\textbf{\textcolor{red}{7856.850}}\\
 Deep Ensemble&Monte-Carlo Dropout&WBF&0.102&0.300&0.372&9267.882\\
  Deep Ensemble &Monte-Carlo Dropout&NMS&0.085&0.231&0.754&8684.015\\
 Deep Ensemble&Monte-Carlo Dropout&Soft NMS&0.069&0.162&0.864&3379.068\\

 \hline
\end{tabular}
\end{adjustbox}
      \caption{BDD100k Ensemble results}
    \label{bdd100k-ensemble}
\end{table}

\section{Discussion}
Deep ensemble networks provide good predictions, but this is always at an increased computational complexity. To alleviate this problem, we proposed EfficientEnsemble which is not only less computationally intensive but produces  results similar to the traditional Deep Ensembles. Tables ~\ref{MOT-16-ensemble}, ~\ref{Kiiti-ensemble}, and ~\ref{bdd100k-ensemble} show the results we reported for all our experiments. In all these tables, metrics reported in bold black represent the performance of the best model. All results in bold red correspond to the results of the second-best models. Considering the results reported in Tables ~\ref{MOT-16-ensemble}, ~\ref{Kiiti-ensemble} and ~\ref{bdd100k-ensemble}, we can see that EfficientEnsemble outperforms its Monte Carlo counterparts on  object detection metrics, ECE and evaluation time. Likewise, we observe that EfficientEnsemble is on par with its traditional counterparts on both object detection metrics and ECE. However, in terms of evaluation time, EfficientEnsemble makes a time saving of approximately 40\%  on all the datasets. Both the traditional method and EfficientEnsemble outperform the baseline models especially in all rows in Tables ~\ref{MOT-16-ensemble},~\ref{Kiiti-ensemble} and ~\ref{bdd100k-ensemble} where WBF is the aggregation function. But, with NMS and Soft NMS, we noticed that the Deep Ensemble networks  perform slightly worse than baseline models sometimes and this is likely because of the fact that NMS and Soft NMS try to remove overlapping boxes which can affect the performance of the network in case the overlapping boxes do not necessarily belong to the same class category. WBF on the other hand uses all the bounding boxes. We decided to discard bounding boxes with very low scores while using WBF. We used an elimination threshold of 0.20 to remove all bounding boxes with scores less than or equal to  0.20. This, therefore, answers the question we posed earlier and we can clearly see that the choice of merging algorithms affects the quality of results.

Also, we discovered that Gaussian YOLOv3 appears to be better calibrated than the baseline Faster R-CNN model. Nevertheless, Faster R-CNN Deep Ensemble models are better calibrated than Gaussian YOLO (see all rows in Tables ~\ref{MOT-16-ensemble}, ~\ref{Kiiti-ensemble} and ~\ref{bdd100k-ensemble} where WTF is the aggregation method).

\begin{figure*}
    \centering
    \includegraphics[scale=0.50]{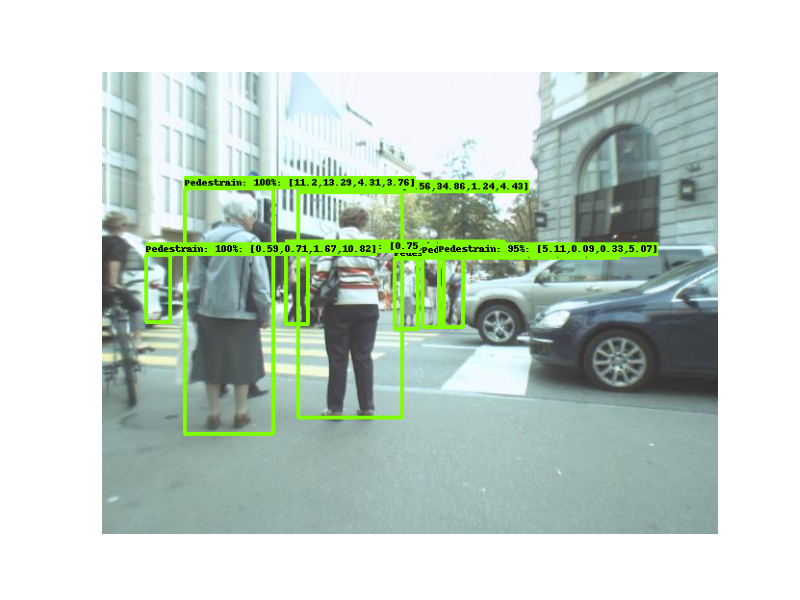}
    \caption{Visualisation of results of EfficientEnsemble on MOT16}
    \label{fig:MOT-results}
\end{figure*}

\begin{figure*}
    \centering
    \includegraphics[scale=0.35]{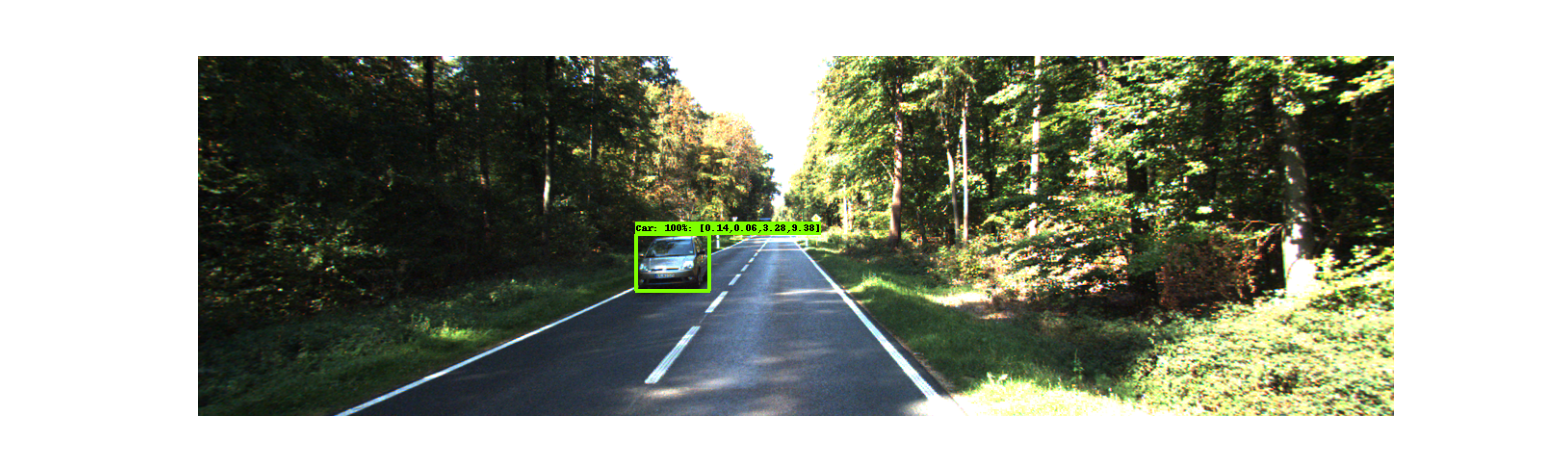}
    \caption{Visualisation of results of EfficientEnsemble on KITTI}
    \label{fig:KITTI-results}
\end{figure*}

We also computed the empirical variance associated with the results of the Deep Ensembles. This is useful as it provides extra information about resulting bounding boxes. Figures ~\ref{fig:MOT-results} and ~\ref{fig:KITTI-results} show the results of EfficientEnsemble on some images in MOT16 and KITTI datasets. We display the scores or confidences and the variance associated with each bounding box. The labels on the images depict the class label, the number in percentage represents the confidence score, and the list of numbers represents the variance associated with the merged box.
\section{Conclusion}
In this paper, we proposed an EfficientEnsemble a lightweight alternative for building Deep Ensembles networks using two-stage object detection models. We conducted extensive experiments to validate our approach and our experimental results show that EfficientEnsemble produces competitive results which are comparable to using $n$ two-stage object detection models in Deep Ensembles. We further observed that better decision fusion algorithms could  boost the performance of an ensemble and produce better-calibrated models.
\section*{Declarations}
\subsection*{Conflict of Interest Statement}
All authors declare that they have no conflicts of interest.
\subsection*{Data Availability Statement}
The data that support the findings of this study are openly available  and can be accessed online using the following links. Datasets : MOT17DET~\footnote{\url{https://motchallenge.net/data/MOT17Det}}, MOT16~\footnote{\url{https://motchallenge.net/data/MOT16/}}, KITTI~\footnote{\url{https://www.cvlibs.net/datasets/kitti/eval_object.php?obj_benchmark=2d}}, and BDDK100k~\footnote{\url{https://doc.bdd100k.com/download.html}}
\bibliography{sn-bibliography}


\end{document}